\documentclass[11pt,a4paper]{article}
\usepackage[hyperref]{naaclhlt2019}
\usepackage{times}
\usepackage{latexsym}
\usepackage{amssymb}
\usepackage{url}
\usepackage{mathtools}
\usepackage{booktabs}
\usepackage{xcolor}
\usepackage{enumitem}
\usepackage[font={footnotesize}]{caption}
\usepackage{algorithm}
\usepackage[noend]{algpseudocode}
\usepackage{float}

\newcommand\ignore[1]{}
\newcommand\jb[1]{ {\color{red} [JB: #1]}}
\newcommand\og[1]{ {\color{purple} [OG: #1]}}
\newcommand\sD{\mathcal{D}}
\newcommand\sY{\mathcal{Y}}
\newcommand\distflip{\textsc{DistFlip}}
\newcommand\distflipfive{\textsc{DistFlip-5}}
\newcommand\distflipten{\textsc{DistFlip-10}}
\newcommand\distflipplus{\textsc{DistFlip+}}

\DeclareMathOperator*{\argmax}{arg\,max}

\aclfinalcopy 
\def\aclpaperid{1265} 

\title{White-to-Black: Efficient Distillation of Black-Box Adversarial Attacks}

\author{Yotam Gil$^{1}$\thanks{\ \ \ Equal contribution}    \and Yoav Chai$^{2}$\footnotemark[1] \and Or Gorodissky$^{1}$\footnotemark[1] \and Jonathan Berant$^{2,3}$ \\
    $^{1}$School of Electrical Engineering, Tel-Aviv University  \\
    $^{2}$School of Computer Science, Tel-Aviv University \\
    $^{3}$Allen Institute for Artificial Intelligence \\
    {\tt \small \{yotamgil@mail, yoavchai1@mail, orarieg@mail, joberant@cs\}.tau.ac.il }}
    
\date{}

\begin{document}

\maketitle

\begin{abstract}
Adversarial examples are important for understanding the behavior of neural models, and can improve their robustness through adversarial training. 
Recent work in natural language processing generated adversarial examples by assuming white-box access to the attacked model, 
and optimizing the input directly against it \cite{ebrahimi:18}. 
In this work, we show that the knowledge implicit in the optimization procedure can be distilled into another more efficient neural network.
We train a model to emulate the behavior of a white-box attack and show that it generalizes well across examples. Moreover, it reduces adversarial example generation time by 19x-39x.
We also show that our approach transfers to a black-box setting, by attacking The Google Perspective API and exposing its vulnerability.
Our attack flips the API-predicted label in 42\% of the generated examples, while humans maintain high-accuracy in predicting the gold label.
\end{abstract}

\section{Introduction}

Adversarial examples \cite{Goodfellow:14} have gained tremendous attention recently, as they elucidate model limitations, and expose vulnerabilities in deployed systems. Work in natural language processing (NLP) either (a) used simple heuristics for generating adversarial examples \cite{jia2017adversarial,belinkov:17,iyyer:18}, or (b) assumed white-box access, where the attacker has access to gradients of the model with respect to the input \cite{feng:18,ebrahimi:18}.
In this approach, adversarial examples are constructed through an optimization process that uses gradient descent to search for input examples that maximally change the predictions of a model. However, developing attacks with only black-box access to a model (no access to gradients) is still under-explored in NLP.

\ignore{
	\begin{table}[t]
		\centering
			\label{Table_0}
			\begin{center}
				\begin{scriptsize}
					\begin{tabular}{c|c|c}
						\toprule
						  \textbf{input} & its an AP article you asshole, how is it not neutral??? & \textcolor{green}{0.96}  	\\
						  \textbf{output}& its an AP article you ass\normalfont{\bf{\underline{n}}}ole, how is it not neutral??? & \textcolor{red}{0.48} 	\\
						\bottomrule
						  \textbf{input}& I think the 1 million sales is total bullshit though. & \textcolor{green}{0.79}	\\
						  \textbf{output}& I think the 1 million sales is total bullsh\normalfont{\bf{\underline{k}}}t though. & \textcolor{red}{0.07} 	\\
						\bottomrule
					\end{tabular}
				\end{scriptsize}
			\end{center}
		\caption{Adversarial examples from our black-box model that flips the toxicity prediction of Google Perspective API. }
		\vspace{-0.35cm}
		\label{tab:adversarial_ex}
	\end{table}
	}

\begin{figure}[t]
  \includegraphics[width=0.48\textwidth]{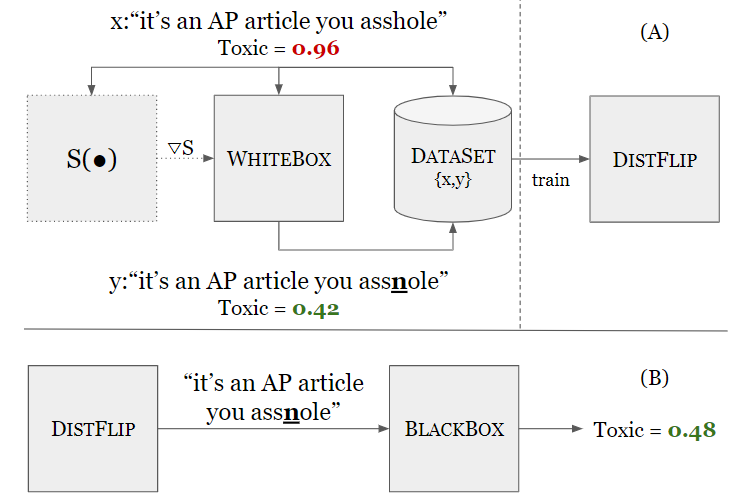}
  \caption{(A) Using a white-box attack we generate adversarial examples for a source toxicity model $S(\cdot)$. We train our black-box attacker, \textsc{DistFlip}, to emulate the white-box attack. (B) We use \textsc{DistFlip} to attack a black-box model.}
  \label{fig:framework}
\end{figure}

Inspired by work in computer vision \cite{papernot2016transferability,liu:17}, we show in this work that a neural network can learn to emulate the optimization process of a white-box attack and generalize well to new examples. Figure~\ref{fig:framework} gives a high-level overview of our approach. We assume a text classification model and a white-box attack that flips characters in the input to modify the model prediction \cite{ebrahimi:18}. We generate output adversarial examples using the white-box attack and train a neural network from these input-output examples to imitate the white-box attack. This results in a much more efficient attack whose run-time is independent of the optimization process. Moreover,  
assuming adversarial examples transfer between different models, our distilled model can now be used to generate adversarial examples for black-box attacks directly.

\ignore{
We focus on \textsc{HotFlip} \cite{ebrahimi:18}, an optimization-based method that flips characters in an input to modify the prediction of a sentence classification model. We generate adversarial examples using \textsc{HotFlip}, and then train our model to emulate the behaviour of \textsc{HotFlip} given only its input-output pairs. This results in a more efficient attack that does not require optimization nor access to gradients. Moreover,  
assuming adversarial examples transfer between different models, the distilled neural network can now be used to generated adversarial examples for black-box attacks directly.
}

We use our approach to attack a toxicity classifier, aimed at detecting toxic language on social media \cite{hosseini2017deceiving}. We show that our model achieves a speed-up of 19x-39x in generating adversarial examples while maintaining similar attack quality, compared to an optimization-based method. We then use our model for a black-box attack against Google Perspective API for detecting toxic sentences, and find that 42\% of our generated sentences are misclassified by the API, while humans agree that the sentences are toxic.

Our code can be downloaded from \url{http://github.com/orgoro/white-2-black}.

\section{Background}
Adversarial examples have been extensively used recently in NLP for probing and understanding neural models \cite{jia2017adversarial,weber:18}. Methods for generating such examples include adding random or heuristically constructed noise \cite{belinkov:17,rodriguez:18,Gao:18}, meaning-preserving modifications that change the surface form \cite{iyyer:18,ribeiro2018semantically}, and human-in-the-loop generation \cite{wallace2018trick}. 
\ignore{
\newcite{Gao:18} used words subtractions for scoring of important words which he later randomly manipulated.
}
A weakness of such models is that they do not directly try to modify the prediction of a model, which can reduce efficacy
\cite{kurakin:16}. In the white-box setting, \newcite{feng:18} have changed the meaning of an input without changing model output using access to gradients, and \newcite{ebrahimi:18} proposed \textsc{HotFlip},
the aforementioned white-box attack that we emulate for flipping input characters.
\ignore{
\jb{what about - \cite{Gao:18}}\og{very relevant to our work he is doing our attention baseline but shows high success perhaps because of the weakness of word based models. Uses a scoring function on words contribution to the classification by subtracting them - then he flips/dups/swaps randomly. His weakness: (1)he works only on word-based - words he manipulates gets unknown symbol (2)changes 10 percents of sentence words and is only weakly optimized because he just randomly flips. each of his passes are done on the attacked model for the scoring}.
}

\ignore{
Deep neural networks (DNN) have gained much popularity in solving NLP problems \citep{sutskever:14}. The increased complexity and the transition from handcrafted features raise serious questions regarding the nature of what exactly these deep networks learn \citep{mitchell:18, feng:18}, and their robustness to noise \citep{belinkov:17, weber:18}. As more algorithms rely on DNNs, so does the importance of making them robust to noisy inputs, and in some sensitive fields to adversarial inputs as well.
}

In computer vision, \newcite{papernot2016transferability} and \newcite{liu:17} have shown that adversarial examples generated by a white-box model can be helpful for a black-box attack. Generating adversarial text examples is more challenging than adversarial images, because images are points in a continuous space, and thus it is easier to apply norm restrictions. Text examples have a discrete structure and thus such approaches have not been investigated for adversarial text generation yet.

\section{\textsc{HotFlip}}
\label{sec:hotflip}

\textsc{HotFlip} \cite{ebrahimi:18} is a white-box method for generating adversarial examples for a character-level neural model. It uses the gradient with respect to a 1-hot input representation to estimate the character flip that incurs the highest cost. We briefly describe \textsc{HotFlip}, which we use to generate training examples for our distilled model.

Let $\mathrm{x} = ((x_1^1, \dots, x_1^n), \dots, (x_m^1, \dots, x_m^n))$ be a sentence represented as a sequence of $m$ characters, encoded as 1-hot vectors over a vocabulary of size $n$. Define $L(\mathrm{x},y)$ to be the loss of a trained model for the input $\mathrm{x}$ with respect to a label $y$. 

\textsc{HotFlip} requires one function evaluation (forward pass) and one gradient computation (backward pass) to compute a first-order estimate of the best possible character flip in $\mathrm{x}$.
Flipping the $i^{\text{th}}$ character from $a$ to $b$ can be represented by this vector:
$
\overrightarrow{v_{i_b}}=(\dots, (0, \dots ,-1, \dots ,1, \dots ,0)_i,  \dots),
$
where $-1$ and $1$ are in the positions for the $a^\text{th}$ and $b^\text{th}$ characters respectively. A first-order estimate of the change in loss can be obtained by computing $\nabla_x L(\mathrm{x},y)$ with back-propagation, and taking a directional derivative along $\overrightarrow{v_{i_b}}$:
\begin{align*}
\nabla_{\overrightarrow{v_{i_b}}} L(\mathrm{x},y)=\nabla_x L(\mathrm{x},y)\cdot \overrightarrow{v_{i_b}}.
\end{align*}

We can now choose the character-flip $a$ to $b$ that maximizes this estimate using the gradients with respect to the input $\mathrm{x}$:
\begin{align*}
\argmax_{i,b}[\nabla L(\mathrm{x},y)\cdot \overrightarrow{v_{i_b}}]= \argmax_{i,b}[\frac{\partial L}{\partial x_{i}^{b}} - \frac{\partial L}{\partial x_{i}^{a}} ].
\end{align*}
To perform a sequence of flips,
any search procedure can be applied.
\textsc{HotFlip} uses beam search of $r$ steps, keeping at each step the top-$K$ flip sequences that increased $L(\mathrm{x},y)$ the most. This require $O(K\cdot r)$ forward and backward passes. Character insertions and deletions are modeled as multiple flips, but for simplicity, we only consider \emph{character flips} in our work.

The main drawbacks of \textsc{HotFlip} are that it does not gain any knowledge from optimizing over multiple examples, and that its efficiency is strongly tied to the search procedure used ($O(K \cdot r)$ forward and backward passes per example for beam-search).
Next, we present our model that overcomes these limitations.

\section{Distilling a Black-box Attack}

We are interested in whether (a) the knowledge in the optimization process of \textsc{HotFlip} can be distilled into a neural model, and (b) whether this model can generalize to a black-box attack. Therefore, the outline of our approach is as follows:
\begin{enumerate}[topsep=0pt,itemsep=0pt,parsep=0pt,wide=0pt,leftmargin=\parindent]
    \item Train a \emph{source text classification model} on data from a similar distribution to the data used to train the \emph{target black-box model}.
    \item Generate adversarial examples by performing white-box optimization (with \textsc{HotFlip}) on the source model.
    \item Train an efficient \emph{attacker} to generate adversarial examples, and perform a black-box attack against the target model.
\end{enumerate}

We assume a training set $\mathcal{D} = \{(\mathrm{x_i}, y_i)\}_{i=1}^N$ used to train a character-based source model $S(\cdot)$ that takes a character sequence $x$ as input, and returns a distribution over the output space $\sY$
(details on the source model are in Section~\ref{sec:experiments}).
We now elaborate on the processes of data generation and training of the attacker.

\paragraph{Data generation}
We take examples $(\mathrm{x},y)$ from $\sD$ and run  \textsc{HotFlip} with search until we obtain an adversarial example $\mathrm{\bar{x}}$ such that the probability of the gold label is low, that is,  $[S(\mathrm{\bar{x}})]_y < \tau$, where $[S(\mathrm{\bar{x}})]_y$ is the probability given by $S(\mathrm{\bar{x}})$ to $y \in \sY$ and $\tau$ is a threshold (we use $\tau=0.15$).

Let $s = ((\mathrm{x^{(0)}} = \mathrm{x}), \mathrm{x^{(1)}}, \dots, (\mathrm{x^{(l)}} = \mathrm{\bar{x}}))$ be the sequence of sentences generating $\mathrm{\bar{x}}$, where every consecutive pair $(\mathrm{x^{(i)}}, \mathrm{x^{(i+1)}})$ differs by a single character:  
the character in position $j^{(i)}$ was flipped to the character $c^{(i)}$. Our attacker is trained from examples $(\mathrm{x^{(i)}}, (j^{(i)}, c^{(i)}))$ generated from 
every pair of consecutive sentences in $s$. For example, if the sentence is the one-word sentence ``Asshole", and after flipping one character it becomes ``Assnole", the example would be (``Asshole", (4, `n')).

\paragraph{Model training}
Our attacker takes a character sequence $\mathrm{x}$ as input and outputs a pair $(j^*, c^*)$, where $j^* \in [1, \dots, m]$ is the position of the character to be flipped, and $c^*$ is the target character.

\begin{figure}[t]
    \center
  \includegraphics[width=0.5\textwidth]{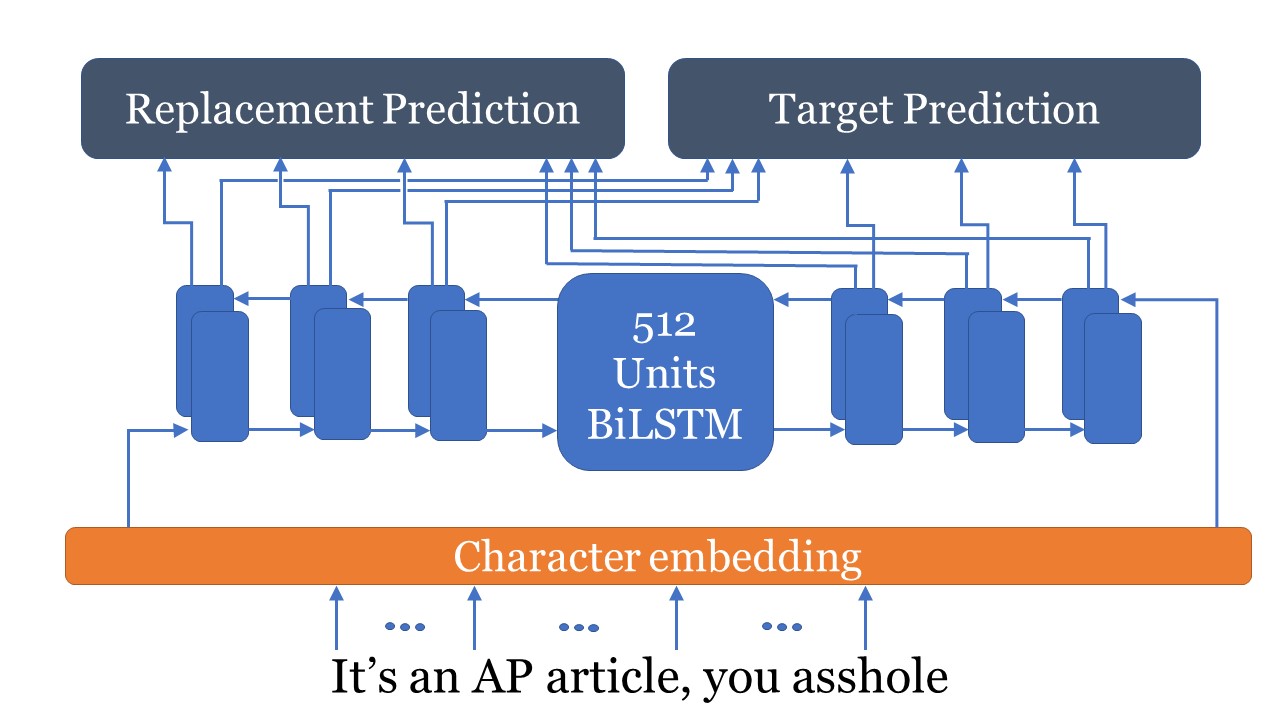}
  \caption{The architecture of our attacker network.}
  \label{fig:architecture}
\end{figure}

Figure~\ref{fig:architecture} describes the architecture of our model. 

Our model embeds each character using a pre-trained 300-dimensional character embedding\footnote{\url{https://github.com/minimaxir/char-embeddings/blob/master/glove.840B.300d-char.txt}}, and then passes the character sequence through a 1-layer bidirectional LSTM \cite{hochreiter1997lstm} with 512-dimensional hidden states. The BiLSTM $h_j$ hidden state in every position are passed through two feed-forward networks, one for replacement prediction (which character to flip) and one for target prediction (what target character to choose). The network that perfoms replacement prediction has 2 hidden layers of dimensions $100$ and $50$ with ReLU activations, and a single logit $l_j$ as output per position. The output distribution over the sentence positions is given by a softmax over all character positions. At inference time $j^*$ is computed with an argmax instead of a softmax.

The network that produces target prediction has two-hidden layers of dimensions $100$ and $100$ with ReLU activations and outputs a vector of logits $v_j \in \mathbb{R}^{96}$ per position with a softmax layer, which provides a distribution over the character vocabulary. The target character $c^*$ is computed at inference time with an argmax over the target position $\mathrm{x_{j^*}}$.

Our loss function is simply the sum of two cross-entropy terms: one for the gold position, and one for the gold character in the gold position.

\ignore{
Our attacker model:
\begin{itemize}\itemsep0em
\item Each sentence was zero-padded or cropped to a fixed size of 500
\item Each character was embedded using a pre-trained embedding of size 300
\item our input layer was fed with the sentences  - $(?, 500, 300)$
\item Each sentence was embedded using a BiLSTM layer $(?, 500, 512)$. 
\item The states ware passed through two classification heads for predicting the replacement character index an target with 
$(?, 500)$ and$(?, 96)$ respectively.
\item Each classification head passed through 2 fully connected layers with ReLU activation.
\item The loss was a weighted some of the two softmax cross-entropy losses computed on the classification heads 
\end{itemize}
}

Running our model is more efficient
that \textsc{HotFlip} and run-time is independent of the optimization procedure. A forward pass in our model is equivalent to $2\cdot K$ steps in \textsc{HotFlip} with beam-search. We show this leads to large practical speed-ups in Section~\ref{sec:experiments}.

\ignore{
One forward pass of our model is equivalent to $2b$ steps of the HotFlip optimization, enabling us to gain an order of magnitude faster results on the average attack length.
}
\section{Experiments}
\label{sec:experiments}

We now empirically investigate whether our method can be used to attack classifiers for 
detecting ``toxic" language on social media. Recently a challenge by Alphabet aimed to improve labeling of toxic comments that are rude and disrespectful.
Alphabet released a dataset\footnote{\url{https://www.kaggle.com/c/jigsaw-toxic-comment-classification-challenge/data}} of 160K comments from Wikipedia discussions, and classified each comment to six labels. We focus on the \emph{Toxic} label only, which  represents 9.5\% of the dataset.

We used the dataset to train the source model $S(\cdot)$, which  runs a 2-layer bidirectional GRU \cite{cho2014gru} over the input $x$, and then uses an attention \cite{bahdanau2015neural} pooling layer to obtain a fixed dimensional vector for $x$. This vector is passed through a feed-forward layer to compute the probability that $x$ is toxic. The accuracy of the source model is 96.5\% AUC -- comparable to the top submissions in the challenge.

\ignore{
We have first implemented simple char-based classifier with a comparable accuracy to the Kaggle challenge top submissions. Choosing  a char-based model rather than a word-based model made good sense on the data, which contain many typos, misspelling and free style text.
}

We used the 13,815 toxic-labeled sentences from the training set to generate adversarial examples for training the attacker as described above.
Dataset generation depends on the search procedure, and we experiment with 3 setups: (a) \textsc{HotFlip-5}: beam-search with $K$=5, (b) \textsc{HotFlip-10}: beam-search with $K$=10, and (C) \textsc{HotFlip+}: a more expensive search procedure that calls $S(\cdot)$ more frequently. We describe the details of this procedure in Appendix~\ref{sec:algorithm}. Because our attacker is independent of the search procedure, inference time is not affected by the search procedure at data generation time. 
This results in three distilled models: \distflipfive{}, \distflipten{}, and \distflipplus{}.\footnote{\distflipfive{} is trained on 64K generated examples, \distflipten{} on 62K examples, and \distflipplus{} on 46K.}

\ignore{
Our next step was to a attack this model in a white-box manner, using HotFlip. We created 72K examples for each of the two configurations - beam search of size 5 and of size 10. We stopped attacking a sentence when the classifier reached a toxicity below 0.15.
Each example in our data set represents one flip in the HotFlip optimization process.

We split our data set to 62K examples used for training, 5K for validation and 7K for test. 
}

\paragraph{Attacking the source model}
\ignore{
We train our models, \distflipfive{} and \distflipten{}, using data from \textsc{HotFlip-5} and \textsc{HotFlip-10} respectively. We first examine the accuracy of our models in predicting the correct target position and character on the test set. Table~\ref{tab:accuracy} shows that our model achieves good performance, finding the correct position with accuracy 65\% and the character target with 96\% accuracy with \distflipten{}.

\begin{table}
\centering
\begin{footnotesize}
\begin{tabular}{l|cccc}
\multicolumn{1}{l}{}                     & \multicolumn{2}{l}{char. position} & \multicolumn{2}{l}{char. target}  \\
\hline
\multicolumn{1}{c|}{beam size} & 5 & 10                                   & 5 & 10                                     \\ 
\hline
top-1 acc.                      & 0.34 & 0.43                                    & 0.67 & 0.75                                      \\ 
\hline
top-5 acc.                       & 0.62 & 0.65                                    & 0.94 & 0.96 
\end{tabular}
\caption{Accuracy of \distflip in predicting the flipped position and target.}
\label{tab:accuracy}
\end{footnotesize}
\end{table}
}

\begin{table*}[t]
		\centering
			\label{Table_1}
			\begin{center}
				\begin{scriptsize}
					\begin{tabular}{c|ccccccccc}
						\toprule

						\ignore{& A & B & C & D& E &F &G &H &I     \\}
						& \distflipplus & \distflipten{} & \distflipfive{} & \textsc{HotFlip+}& \textsc{HotFlip-10} &\textsc{HotFlip-5} &\textsc{HotFlip-1} & \textsc{Rand.} & \textsc{Att.}     \\
						\midrule
						\bf{\#flips} & 5.05& 7.5 & 8.6 & 3.2& 4.5 & 4.6 & 13.8 & 56.0 & 70.6 \\
						\bf{\#flips for 85\%} & 1.88& 2.27 & 2.30 & 1.66& 2.16 & 2.18 & 2.24 & 24.57 & 36.09 \\
						\bf{Flip slow-down} & 1x& 1x & 1x & 168.8x & 43.3x & 21.3x & 6.1x & - & - \\
						\bf{Attack slow-down} & 1x& 1.48x & 1.7x & 108x & \textbf{38.6x} & \textbf{19.4x} & 16.7x & - & - \\
						\bottomrule
					\end{tabular}
					\caption{
					Average number of flips to change the prediction of toxic sentences, average number of flips to change the prediction for 85\% of the examples that do not contain repeated profanities, slow-down per flip compared to \distflipplus, and slow-down per sentence attack compared to \distflipplus. One character-flip using \distflip{} takes 12ms on an Nvidia GTX1080Ti. }
					\label{tab:survival_rate}
				\end{scriptsize}
			\end{center}
\end{table*}

\ignore{
\begin{table*}[t]
		\centering
			\label{Table_1}
			\begin{center}
				\begin{scriptsize}
					\begin{tabular}{c|cccccccc}
						\toprule

						\ignore{& A & B & C & D& E &F &G &H &I     \\}
						& \distflipplus & \distflipten{} & \distflipfive{} & \textsc{HotFlip+}& \textsc{HotFlip-10} &\textsc{HotFlip-5} & \textsc{R.} & \textsc{A.}     \\
						\midrule
						\bf{\#flips} & 5.05& 7.5 & 8.6 & \textbf{3.2}& 4.5 & 4.6 & 56.0 & 70.6 \\
						\bf{\#flips for 85\%} & 1.88& 2.27 & 2.30 & \textbf{1.66}& 2.16 & 2.18  & 24.57 & 36.09 \\
						\bf{Flip slow-down} & 1x& 1x & 1x & 168.8x & 43.3x & 21.3x  & - & - \\
						\bf{Attack slow-down} & 1x& 1.48x & 1.7x & 108x & 38.6x & 19.4x & - & - \\
						\bottomrule
					\end{tabular}
					\caption{
					Average number of flips needed to change the prediction of toxic sentences, average number of character-flips needed to change the prediction for 85\% of the examples that are easier, slow-down per flip compared to \distflipplus, and slow-down per sentence attack compared to \distflipplus. \textsc{R.} and \textsc{A.} correspond to the \textsc{Random} and \textsc{Attention} baselines.}
					\label{tab:survival_rate}
				\end{scriptsize}
			\end{center}
\end{table*}
}

We compare the performance of \distflip{} variants
against 
\textsc{HotFlip} variant, including \textsc{HotFlip-1} ($K$=1).
We also compare to a \textsc{Random} baseline, which chooses a position and target character randomly, and to an \textsc{Attention} baseline, which uses the attention layer of $S(\cdot)$ to choose the character position with maximum attention to flip, and selects a target character randomly.

Table~\ref{tab:survival_rate} summarizes our results. We report the average number of flips required to change the prediction of toxic sentences in the source model, the slow-down per single character-flip, and the slow-down per attack, which is computed by multiplying slow-down per flip by the ratio of the number of flips required per attack. 
Because roughly 15\% of the examples in the dataset mostly contain repeated profanities that require many flips,
we also report the average number of flips for the other 85\% of the examples.

We observe that \textsc{HotFlip+} requires the fewest flips to change model prediction, but attacks are very slow. 
The number of flips per attack for \distflipplus{} is comparable to \textsc{HotFlip-5} and \textsc{HotFlip-10}, but it achieves a speed-up of 19x-39x. \distflipfive{} and \distflipten{} require a few more flips compared to \distflipplus{}. Overall attack quality is high, with less than two flips necessary for 85\% of the examples for \distflipplus{}. 

\begin{figure}[t]
  \includegraphics[width=0.5\textwidth]{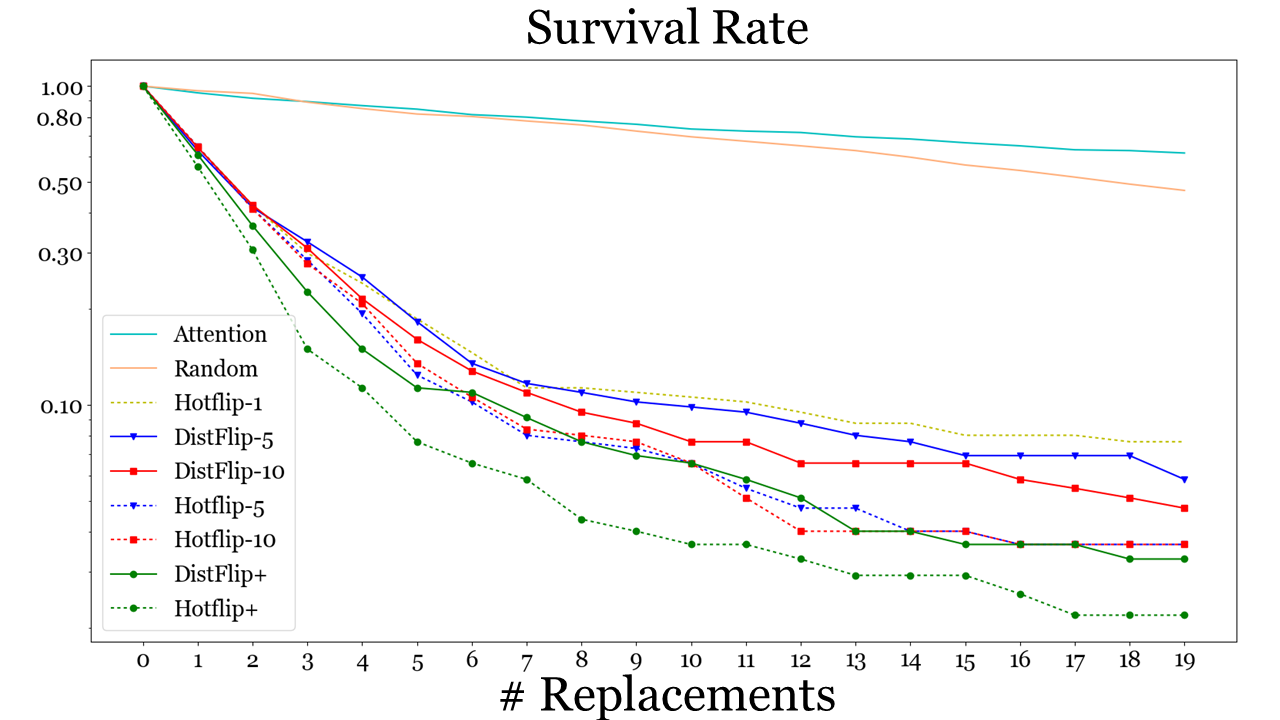}
  \caption{Proportion of sentences classified as toxic as a function of the number of flips for all models on the test set.}
  \label{fig:survival_rate}
\end{figure}

Figure~\ref{fig:survival_rate} 
provides a more fine-grained view of the results by
showing the proportion of sentences classified as toxic as a function of the number of flips. Overall, the picture is similar to Table~\ref{tab:survival_rate}, with \distflipplus{} being comparable to \textsc{HotFlip-10}.

\paragraph{Attacking The Google Perspective API}
The Google perspective API\footnote{\url{https://www.perspectiveapi.com}} returns the probability that a sentence is toxic, where probability $>0.7$ is classified as \emph{toxic}, $<0.3$ is \emph{non-toxic}, and otherwise \emph{uncertain}. The model itself is not publicly available. We randomly selected 136 toxic examples from the validation set and attacked them with \distflipplus{} until the source model probability was $<0.5$. 

We measured the toxicity probability before and after our attack and found that the average toxicity probability decreased from $0.9$ to $0.67$, with an average of 5.0 flips per sentence. The label is flipped from \emph{toxic} to \emph{uncertain} or \emph{non-toxic} in 42\% of these examples.

\ignore{
Google perspective API is meant to automation flag of toxic comments. The API returns a probability score between 0 to 1, with higher values indicating greater likelihood of toxicity existence. The model itself is not publicly available. We have used this API on randomly selected sentences to validate our attack ability to generalize and be used as a black box attack.

we randomly chose 135 sentences from our test set and attacked it using our model until reaching toxicity score of below 0.5. We evaluated toxicity with Google Perspective before and after our attack. Out of the 135 sentences, Google couldn't evaluate the toxicity for 8 resulting sentences. For the rest of the sentences, we were able to generalize and reduce the toxicity probability in Google's unseen model, as shown in \ref{Table_3}.

	\begin{table}[th]
		\centering
			\label{Table_3}
			\begin{center}
				\begin{small}
					\begin{tabular}{c|cccc}
						\toprule

						Model  & Toxicity  & Toxicity  & Delta \\
						        &Before Attack&After Attack&  \\
						\midrule
						  \bf{Google}  & 0.9& 0.67& 0.23
						\\
						\bottomrule
					\end{tabular}
				\end{small}
			\end{center}
		\caption{Average Toxicity of 127 sentences before and after the attack. The average toxicity decrease in Google Perspective was 0.23. Examples of attacked sentences can be found in the Appendix Table ~\ref{Table_5}. }
		\vspace{-0.55cm}
	\end{table}
	}

	\begin{table*}[t!]
		\centering
			\begin{center}
				\begin{footnotesize}
				    \hspace*{-1cm}
					\begin{tabular}{llc}
						\toprule

						\textbf{Original}  & \textbf{Adversarial}  & \textbf{Google score}  \\
						\midrule
						 decline=I don't fucking think so
                        & decline=I don't \normalfont{\bf{v}}ucking think so
                        & 0.79 $\rightarrow$ 0.07\\
                        
                        its an AP article you asshole, how is it not neutral???
                        & 
                        its an AP article you ass\normalfont{\bf{n}}ole, how is it not neutral???

                        & 0.96 $\rightarrow$ 0.48\\
                        
                        I think the 1 million sales is total bullshit though. 
                        & 
                        I think the 1 million sales is total bullsh\normalfont{\bf{k}}t though.
                        
                        & 0.79 $\rightarrow$ 0.07\\
                        
                        do not know juden! silence is golden juden, now shut up.
                        & 
                        do not know juden! silence is golden juden, now \normalfont{\bf{x}}h\normalfont{\bf{j}}t up.
                        
                        & 0.69 $\rightarrow$ 0.37\\
                        
                        This article sucks . . woo woo wooooooo
                        & 
                        This article \normalfont{\bf{x}}u\normalfont{\bf{x}}ks . . woo woo wooooooo
                        & 0.93 $\rightarrow$ 0.22\\
                        
                       Also bring back the brendle article you piece of shit.& 
                       Also bring back the brendle article you p\normalfont{\bf{k}}ece of \normalfont{\bf{x}}h\normalfont{\bf{k}}t.

                        & 0.98 $\rightarrow$ 0.32\\

                        to be driven away and die& 
                        to be driven away and d\normalfont{\bf{k}}e
                        & 0.82 $\rightarrow$ 0.32\\

						\bottomrule
					\end{tabular}\hspace*{-1cm}
					\caption{Examples of sentences attacked by \distflipten{} and The Google Perspective toxicity score before and after the attack.}
			        \label{tab:examples}
				\end{footnotesize}
				
			\end{center}
	\end{table*}

\paragraph{Human validation}
To validate that toxic sentences remain toxic after our attack, we showed 5 independent annotators a total of 150 sentences from three classes: toxic sentences, non-toxic sentences, and attacked sentences (attacked by \distflipfive{}). The same annotator was never shown a toxic sentence and its attacked counterpart.
We asked annotators whether sentences are toxic, and measured average annotated toxicity.

We found that 89.8\% of the toxic sentences were annotated as toxic, compared to 87.6\% in the attacked toxic sentences. This shows that humans clearly view the sentences as toxic, even after our attack.
Table~\ref{tab:examples} shows examples for sentences attacked by \distflipten{} and the change in toxicity score according to The Google Perspective API.

\ignore{
\begin{table}[t]
		\centering
			\begin{center}
				\begin{footnotesize}
					\begin{tabular}{c|ccc}
						\toprule
						Class  & Non-Tox.  & Tox.  & Attacked Tox.\\
						\midrule
						  \bf{Toxicity}  & 0.149\pm{0.23}&  0.898\pm{0.17}&  0.876\pm{0.20}
						\\
						\bottomrule
					\end{tabular}
				\end{footnotesize}
			\end{center}
		\caption{Average toxicity and standard deviation for toxic sentences, non-toxic sentences, and attacked sentences.}
        \label{tab:human}
	\end{table}
}
\ignore{
The limit of the amount of characters changed is a good regularize of the attack magnitude but it doesn't ensure that the output sentence remain in the same class i.e. toxic or non-toxic. We used crowd sourcing to evaluate the class of the attacked sentences.

We evaluated a total of 150 sentences consisting of: non-toxic sentences, toxic original sentences and toxic sentences attacked by our method. Each sentence was tagged by 5 independent taggers with the restriction that no tagger is exposed to two different versions of the same original sentence. The results are summarized in Table ~\ref{Table_4}.
While suffering a some decrease in readability most of the attacked sentences maintained their original class toxic showing that indeed the attacked sentence is an adversarial example for attacker. 
}

\section{Conclusion}
In this work we present a general approach for distilling the knowledge in any white-box attack into a trained model. We show that this results in substantial speed-up (19x-39x) while maintaining comparable quality. Moreover, we show that our attack transfers to a black-box setting:  we expose the vulnerability of The Google Perspective API, and are able to change the prediction for 42\% of input toxic examples, while humans easily detect that examples are still toxic.

\section*{Acknowledgements}
We thank Raja Giryes for supporting this research, and the anonymous reviewers for their constructive feedback. 
This research was partially supported by The Israel Science Foundation grant 942/16, the Blavatnik Computer Science Research Fund, and The
Yandex Initiative for Machine Learning.

\bibliography{naaclhlt2019}

\begin{thebibliography}{18}
\expandafter\ifx\csname natexlab\endcsname\relax\def\natexlab#1{#1}\fi

\bibitem[{Bahdanau et~al.(2015)Bahdanau, Cho, and Bengio}]{bahdanau2015neural}
D.~Bahdanau, K.~Cho, and Y.~Bengio. 2015.
\newblock Neural machine translation by jointly learning to align and
  translate.
\newblock In \emph{International Conference on Learning Representations
  (ICLR)}.

\bibitem[{Belinkov and Bisk(2017)}]{belinkov:17}
Yonatan Belinkov and Yonatan Bisk. 2017.
\newblock Synthetic and natural noise both break neural machine translation.
\newblock \emph{arXiv preprint arXiv:1711.02173}.

\bibitem[{Cho et~al.(2014)Cho, van Merri{\"e}nboer, Bahdanau, and
  Bengio}]{cho2014gru}
K.~Cho, B.~van Merri{\"e}nboer, D.~Bahdanau, and Y.~Bengio. 2014.
\newblock On the properties of neural machine translation: Encoder-decoder
  approaches.
\newblock \emph{arXiv preprint arXiv:1409.1259}.

\bibitem[{Ebrahimi et~al.(2018)Ebrahimi, Rao, Lowd, and Dou}]{ebrahimi:18}
Javid Ebrahimi, Anyi Rao, Daniel Lowd, and Dejing Dou. 2018.
\newblock Hotflip: White-box adversarial examples for text classification.
\newblock In \emph{Proceedings of the 56th Annual Meeting of the Association
  for Computational Linguistics (Volume 2: Short Papers)}, volume~2, pages
  31--36.

\bibitem[{Feng et~al.(2018)Feng, Wallace, II, Iyyer, Rodriguez, and
  Boyd-Graber}]{feng:18}
Shi Feng, Eric Wallace, Alvin~Grissom II, Mohit Iyyer, Pedro Rodriguez, and
  Jordan Boyd-Graber. 2018.
\newblock \href {http://arxiv.org/abs/1804.07781} {Pathologies of neural models
  make interpretations difficult}.

\bibitem[{Gao et~al.(2018)Gao, Lanchantin, Soffa, and Qi}]{Gao:18}
Ji~Gao, Jack Lanchantin, Mary~Lou Soffa, and Yanjun Qi. 2018.
\newblock \href {https://doi.org/10.1109/spw.2018.00016} {Black-box generation
  of adversarial text sequences to evade deep learning classifiers}.
\newblock \emph{2018 IEEE Security and Privacy Workshops (SPW)}.

\bibitem[{Goodfellow et~al.(2014)Goodfellow, Shlens, and
  Szegedy}]{Goodfellow:14}
Ian~J. Goodfellow, Jonathon Shlens, and Christian Szegedy. 2014.
\newblock Explaining and harnessing adversarial examples.
\newblock \emph{CoRR}, abs/1412.6572.

\bibitem[{Hochreiter and Schmidhuber(1997)}]{hochreiter1997lstm}
S.~Hochreiter and J.~Schmidhuber. 1997.
\newblock Long short-term memory.
\newblock \emph{Neural Computation}, 9(8):1735--1780.

\bibitem[{Hosseini et~al.(2017)Hosseini, Kannan, Zhang, and
  Poovendran}]{hosseini2017deceiving}
Hossein Hosseini, Sreeram Kannan, Baosen Zhang, and Radha Poovendran. 2017.
\newblock Deceiving google's perspective api built for detecting toxic
  comments.
\newblock \emph{The Bright and Dark Sides of Computer Vision: Challenges and
  Opportunities for Privacy and Security workshop@CVPR}.

\bibitem[{Iyyer et~al.(2018)Iyyer, Wieting, Gimpel, and Zettlemoyer}]{iyyer:18}
Mohit Iyyer, John Wieting, Kevin Gimpel, and Luke Zettlemoyer. 2018.
\newblock Adversarial example generation with syntactically controlled
  paraphrase networks.
\newblock \emph{NAACL}.

\bibitem[{Jia and Liang(2017)}]{jia2017adversarial}
Robin Jia and Percy Liang. 2017.
\newblock Adversarial examples for evaluating reading comprehension systems.
\newblock \emph{EMNLP}.

\bibitem[{Kurakin et~al.(2016)Kurakin, Goodfellow, and Bengio}]{kurakin:16}
Alexey Kurakin, Ian Goodfellow, and Samy Bengio. 2016.
\newblock Adversarial examples in the physical world.
\newblock \emph{ICLR workshop}.

\bibitem[{Liu et~al.(2017)Liu, Chen, Liu, and Song}]{liu:17}
Yanpei Liu, Xinyun Chen, Chang Liu, and Dawn Song. 2017.
\newblock Delving into transferable adversarial examples and black-box attacks.
\newblock \emph{ICLR}.

\bibitem[{Papernot et~al.(2016)Papernot, McDaniel, and
  Goodfellow}]{papernot2016transferability}
Nicolas Papernot, Patrick McDaniel, and Ian Goodfellow. 2016.
\newblock Transferability in machine learning: from phenomena to black-box
  attacks using adversarial samples.
\newblock \emph{arXiv preprint arXiv:1605.07277}.

\bibitem[{Ribeiro et~al.(2018)Ribeiro, Singh, and
  Guestrin}]{ribeiro2018semantically}
Marco~Tulio Ribeiro, Sameer Singh, and Carlos Guestrin. 2018.
\newblock Semantically equivalent adversarial rules for debugging nlp models.
\newblock In \emph{Proceedings of the 56th Annual Meeting of the Association
  for Computational Linguistics (Volume 1: Long Papers)}, volume~1, pages
  856--865.

\bibitem[{Rodriguez and Rojas-Galeano(2018)}]{rodriguez:18}
Nestor Rodriguez and Sergio Rojas-Galeano. 2018.
\newblock \href {http://arxiv.org/abs/1801.01828} {Shielding google's language
  toxicity model against adversarial attacks}.

\bibitem[{Wallace et~al.(2018)Wallace, Rodriguez, Feng, and
  Boyd-Graber}]{wallace2018trick}
Eric Wallace, Pedro Rodriguez, Shi Feng, and Jordan Boyd-Graber. 2018.
\newblock Trick me if you can: Adversarial writing of trivia challenge
  questions.
\newblock \emph{ACL Student Research Workshop}.

\bibitem[{Weber et~al.(2018)Weber, Shekhar, and Balasubramanian}]{weber:18}
Noah Weber, Leena Shekhar, and Niranjan Balasubramanian. 2018.
\newblock The fine line between linguistic generalization and failure in
  seq2seq-attention models.
\newblock \emph{arXiv preprint arXiv:1805.01445}.

\end{thebibliography}
\bibliographystyle{acl_natbib}

\clearpage 
\appendix
\onecolumn
\section{Appendix: HotFlip+}
\label{sec:algorithm}

Algorithm \ref{alg:hotflip+} describes the search procedure of \textsc{HotFlip+}. The main motivation behind this search procedure is that \textsc{HotFlip} uses gradients to estimate the character-flip
that maximally changes the predictions of a model,
but this estimate is not guaranteed to be correct.
In \textsc{HotFlip+} we try to overcome this limitation by performing pruning with gradients, and then actually evaluating a larger number of possible flips by running the source model on many possible flips. This makes the search procedure slower, because we have to run the source model (forward pass) much more frequently. For this algorithm we use beam size $3$.

In Algorithm \ref{alg:hotflip+}, the \emph{toxicity score} of a sentence is the result of running it through the source model, and the \emph{beam score} is the first-order estimate described in Section~\ref{sec:hotflip}.

\begin{algorithm*}[h]
    \caption{HotFlip+ }\label{alg:hotflip+}
    \begin{algorithmic}[1]
    \Procedure{\textsc{HotFlip+}}{sentence}
    \State beam $\leftarrow$ Initialize beam with the original sentence and its toxicity score. 
    \While{True}
    \State $bf, tox \leftarrow$ Pop from the beam the flipped sentence with lowest toxicity, and its toxicity.

	\If{$tox < 0.5$}
	    \State $break$
	\EndIf
	\State create a new beam.

    \For{every beam entry \textbf{in} the current beam} 
    \State compute all possible flipped sentences and their beam score (as in \textsc{HotFlip}).

    \For{$flip\_sent,flip\_score$ \textbf{in} flipped sentences}

    \State $min\_score \leftarrow$ minimal beam score on the beam.

    \If{$flip\_score > min\_score$} \Comment{Prune using beam score}
        \State $tox \leftarrow $ Compute the toxicity of $flip\_sent$ with a forward pass.
		\State $max\_tox\_in\_beam \leftarrow $ Pop from the new beam the maximal toxicity score.
        \If{ $tox < max\_tox\_in\_beam$ } 
            \State push $flip\_sent, flip\_score, tox$ to the new beam.
        \EndIf
	\EndIf
    \EndFor{}
    \EndFor{}
    \EndWhile\label{}
    \State \textbf{return} $bf$
    \EndProcedure
    \end{algorithmic}
\end{algorithm*}

\end{document}